\documentclass[conference]{IEEEtran}
\usepackage{amsmath,amsfonts}
\usepackage[caption=false,font=normalsize,labelfont=sf,textfont=sf]{subfig}
\usepackage{textcomp}

\usepackage[utf8]{inputenc}
\usepackage[english]{babel}

\usepackage{todonotes}
\usepackage[acronym]{glossaries}
\glsenableentrycount
\usepackage[commandnameprefix=always]{changes} 

\usepackage[hidelinks]{hyperref}
\usepackage{cleveref}
\usepackage{babel}
\usepackage{url}
\usepackage[perpage]{footmisc}

\usepackage{graphicx}
\usepackage[binary-units]{siunitx}
\usepackage{overpic}

\usepackage{ifdraft}
\usepackage{booktabs}

\usepackage[normalem]{ulem}
\usepackage{scalerel}
\usepackage[switch]{lineno}

\usepackage{subfiles}
\usetikzlibrary{calc}

\usepackage[backend=biber,maxbibnames=1,minbibnames=1,mincitenames=1,firstinits=true,sorting=none,style=ieee,doi=true,eprint=false,isbn=false,url=false]{biblatex}
\newcommand\newreplacement[2]{\def#1/{#2}}
\newreplacement{\BSS}{BrainScaleS}
\newreplacement{\BSSS}{BrainScaleS system}
\newreplacement{\BSSWSS}{BrainScaleS wafer-scale system}
\newreplacement{\BSSWM}{BrainScaleS wafer module}

\newacronym{bss1}{\mbox{BSS-1}}{Brain\mbox{ScaleS-1}}
\newacronym{bss2}{\mbox{BSS-2}}{Brain\mbox{ScaleS-2}}
\newacronym{fpga}{FPGA}{field-programmable gate array}
\newacronym{fg}{FG}{Single-Poly Floating Gate}
\newacronym{hicann}{HICANN}{High Input Count Analog Neural Network}
\newacronym{psp}{PSP}{post synaptic potential}
\newacronym{isi}{ISI}{inter spike interval}
\newacronym{sfc}{synfire chain}{synchronous firing chain}
\newacronym{wsi}{WSI}{wafer-scale integration}
\newacronym{adex}{AdEx}{adaptive exponential integrate-and-fire}
\newacronym{lif}{LIF}{leaky integrate-and-fire}
\newacronym{asic}{ASIC}{application-specific integrated circuit}
\newacronym{pcb}{PCB}{printed circuit board}
\newacronym{dsl}{DSL}{domain-specific language}
\newacronym{api}{API}{application programming interface}

\usepackage{changes}

\begin{document}

\title{Demonstrating the Advantages of Analog Wafer-Scale Neuromorphic Hardware}

\author{%
	\IEEEauthorblockN{%
		Hartmut Schmidt\IEEEauthorrefmark{1}\IEEEauthorrefmark{3}, Andreas Grübl\IEEEauthorrefmark{1}, Jos{\'e} Montes\IEEEauthorrefmark{1},\\
		Eric Müller\IEEEauthorrefmark{1}, Sebastian Schmitt\IEEEauthorrefmark{2} and Johannes Schemmel\IEEEauthorrefmark{1}}%
	\IEEEauthorblockA{\IEEEauthorrefmark{1}\textit{Kirchhoff Institute for Physics}, Heidelberg University, Germany}%
	\IEEEauthorblockA{\IEEEauthorrefmark{2}Göttingen, Germany}%
	\IEEEauthorblockA{\IEEEauthorrefmark{3}\href{mailto:hschmidt@kip.uni-heidelberg.de}{hschmidt@kip.uni-heidelberg.de}}%
}

\maketitle

\begin{abstract}
	As numerical simulations grow in size and complexity, they become increasingly resource-intensive in terms of time and energy.
While specialized hardware accelerators often provide order-of-magnitude gains and are state of the art in other scientific fields, their availability and applicability in computational neuroscience is still limited.
In this field, neuromorphic accelerators, particularly mixed-signal architectures like the BrainScaleS systems, offer the most significant performance benefits.
These systems maintain a constant, accelerated emulation speed independent of network model and size.
This is especially beneficial when traditional simulators reach their limits, such as when modeling complex neuron dynamics, incorporating plasticity mechanisms, or running long or repetitive experiments.
However, the analog nature of these systems introduces new challenges.
In this paper we demonstrate the capabilities and advantages of the BrainScaleS-1 system and how it can be used in combination with conventional software simulations.
We report the emulation time and energy consumption for two biologically inspired networks adapted to the neuromorphic hardware substrate: a balanced random network based on \citeauthor{brunel_jcns2000} and the cortical microcircuit from \citeauthor{potjans2012cell}.

\end{abstract}


\section{Introduction}

The numerical simulation of the dynamic evolution of large-scale spiking neural networks remains a significant challenge due to the massively parallel interactions of their neurons and synapses, even on the fastest numerical accelerators~\cite{kurth2022sub}.
While simulations of point-type neurons are often communication-bound~\cite{jordan2018extremely}, the introduction of complex node-local computations---such as in structured neurons or for certain types of plasticity---can yield a computation-bound behavior, e.g.,~\cite{kumbhar2019coreneuron,herten2024application}.
Some neuromorphic substrates present a potential solution to this challenge, offering the capacity for inherently parallel computation of neuron and synapse models while enabling spike-based communication.
In recent years, hybrid forms of neuromorphic systems have been developed; see~\cite{thakur2018mimicthebrain_nourl} for a systems overview.
One such platform is the \gls{bss1} system.
Through its analog, wafer-scale architecture, it supports high neuron and synapse counts and enables highly accelerated emulations.
However, this gain in performance comes at the cost of reduced flexibility in terms of model selection and network structure, which can pose challenges for experimental execution.

In this work, we demonstrate this system's potential through the emulation of two large-scale biologically inspired models:
the balanced random network, as proposed by~\cite{brunel_jcns2000}, and the cortical microcircuit, as outlined by~\cite{potjans2012cell}.
Both models are adapted in conventional software simulation to align with the constraints of the neuromorphic hardware before being emulated on it.
Based on our results, we evaluate the hardware's emulation performance and discuss the advantages and practical applications of analog neuromorphic systems, particularly in combination with conventional numeric simulators.

\section{The BrainScaleS-1 System}

In this work, we utilize the \gls{bss1} wafer-scale neuromorphic architecture as an accelerator for emulating biologically-inspired spiking neural networks.
Built in \SI{180}{\nano\meter} CMOS technology, \gls{bss1} employs analog circuits for neurons and synapses with spike events transmitted digitally.
This physical emulation of network dynamics eliminates the need for solving differential equations, enabling operation in continuous time.
Additionally, configurable neuron parameters allow for adjustable acceleration factors, typically \num{10000} times faster than biological real-time, so that \num{1} second of biological activity can be emulated in \num{0.1} milliseconds.

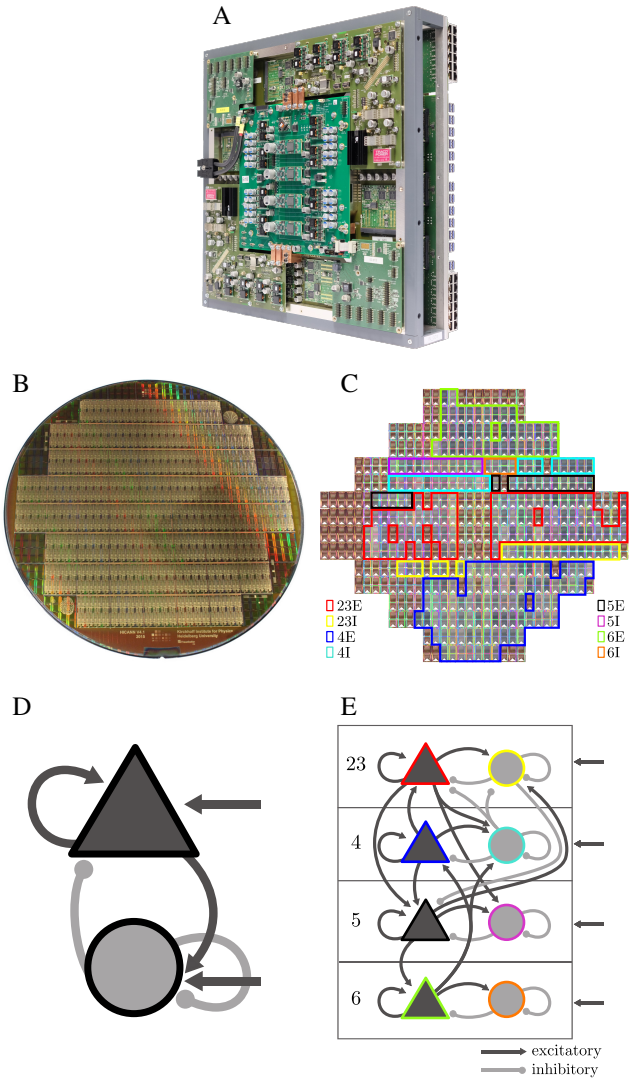
\begin{figure}
	\centering
	\begin{tikzpicture}
		\tikzset{
			panel/.style={
				inner sep=0pt, outer sep=0, execute at begin node={\tikzset{anchor=center, inner sep=.33333em}}
			},
			label/.style={
				anchor=north west, inner sep=0, outer sep=0
			}
		}
		\node[panel, anchor=south west] (a) at (0.3\columnwidth,  0) {
			\includegraphics[width=.39\columnwidth]{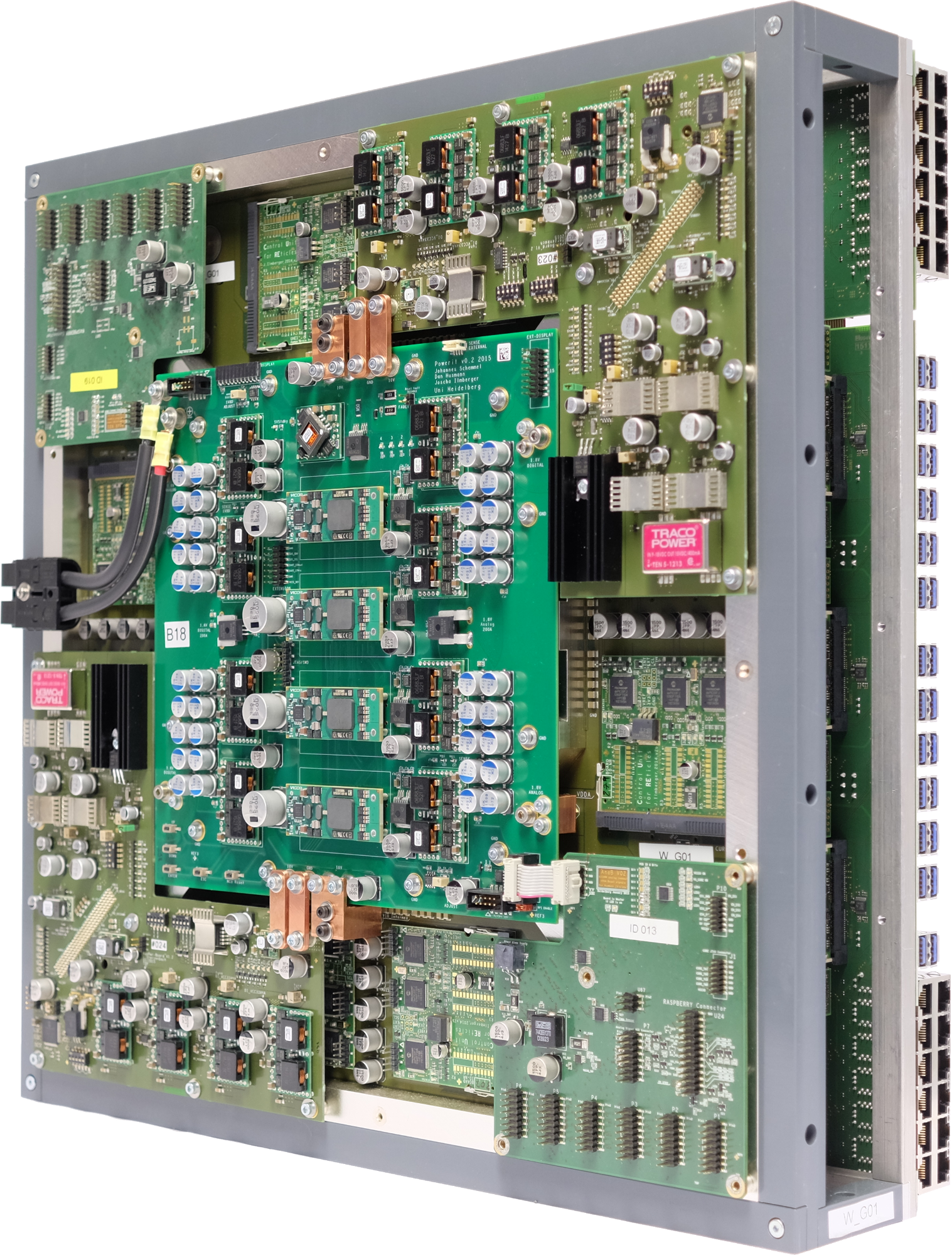}
		};
		\node[label] (A) at ($(a.north west) + (0.2, 0)$) {\textcolor{black}{A}};
		\node[panel, anchor=south west] (b) at (0,  -4.2) {
			\includegraphics[width=.47\columnwidth]{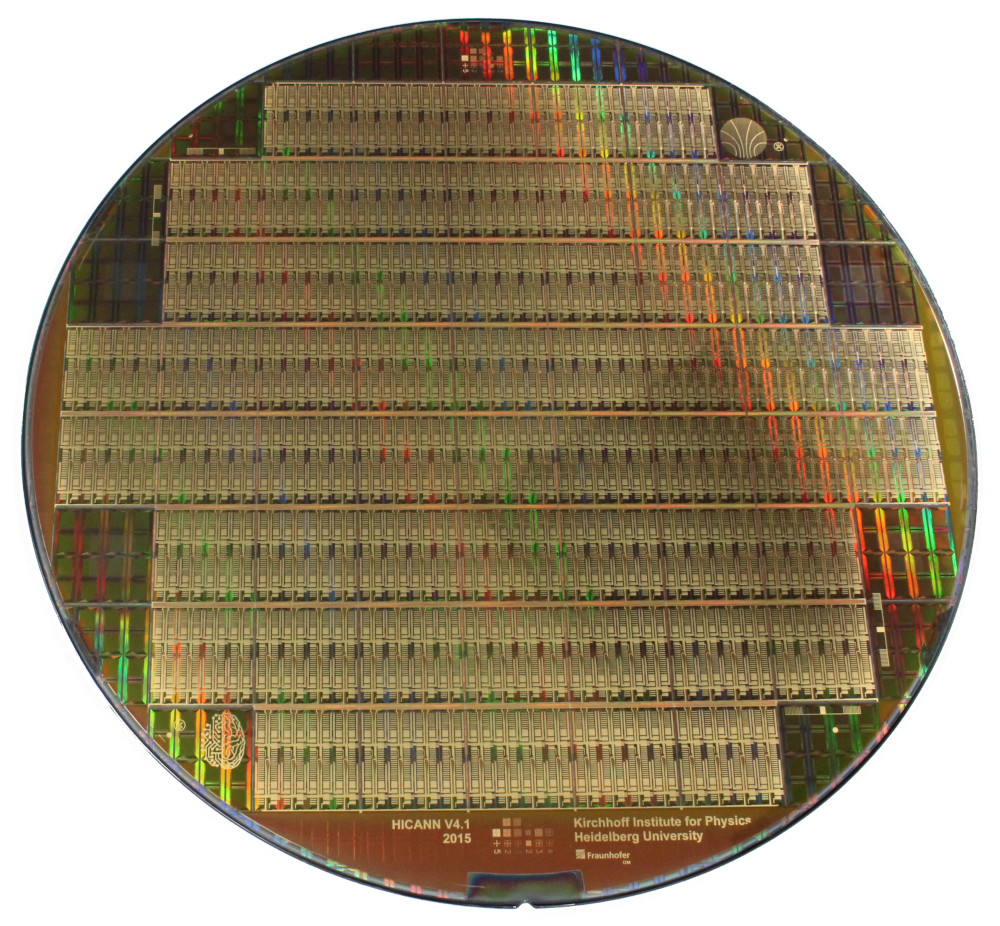}
		};
		\node[label] (B) at ($(b.north west) + (0.2, 0)$) {\textcolor{black}{B}};
		\node[panel, anchor=south west] (c) at (.48\columnwidth, -4.2) {
			\includegraphics[width=.47\columnwidth]{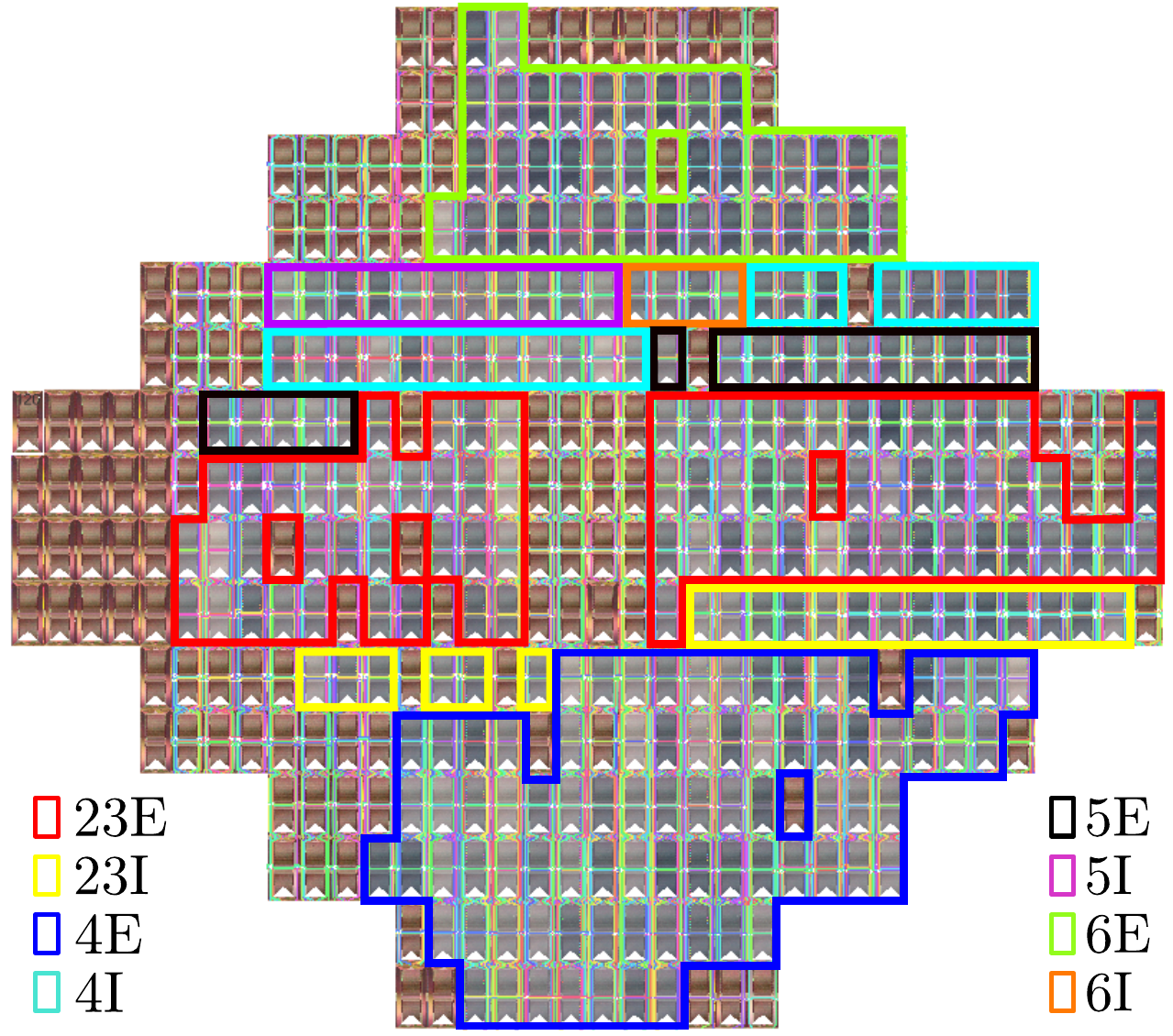}
		};
		\node[label] (C) at ($(b.north east) + (0.4, 0)$) {C};
		\node[panel, anchor=south west] (d) at (0.4, -8.9) {
			\includegraphics[width=.35\columnwidth]{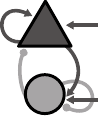}
		};
		\node[label] (D) at ($(b.north west) + (0.2, -4.3)$) {D};
		\node[panel, anchor=south west] (e) at (.51\columnwidth, -9.7) {
			\includegraphics[width=.4\columnwidth]{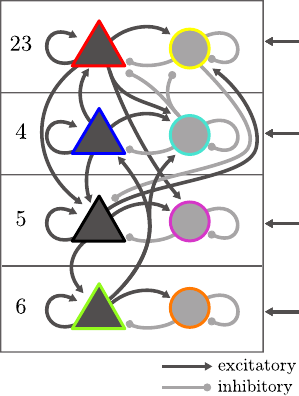}
		};
		\node[label] (E) at ($(b.north east) + (0.4, -4.3)$) {E};
	\end{tikzpicture}
	\caption{%
		A:
		Photograph of a fully assembled \gls{bss1} system.
		B:
		Close up of a \gls{bss1} wafer positioned at the center of the system shown in A.
		Wafer-scale integration is achieved through a redistribution layer applied atop the wafer.
		C:
		Visualization of the adapted cortical microcircuit mapped to a \gls{bss1} wafer.
		Each \gls{asic} on the wafer is depicted as a rectangle with a white triangle at the bottom.
		The neuron placement is represented by the use of blue coloration, with darker shades indicating higher neuron counts.
		Connections are visualized as colored lines routed along the edges of the \glspl{asic}.
		The positions of different model populations (see E) on the wafer are highlighted with distinct colored borders.
		Due to design constraints, not all \glspl{asic} at the wafer's edge, as shown in B, are available for mapping.
		D:
		Connectivity of the balanced random network model.
		E:
		Structure of the cortical microcircuit model.
		Only connections with probabilities greater than \num{0.04} are displayed.
		\label{fig:hardware_and_mrvisu}
	}
\end{figure}

Making use of wafer-scale integration~\cite{zoschkeguettler2017rdlembedding}, \gls{bss1} incorporates \num{384} \glspl{asic} on a \SI{20}{\centi\meter} silicon wafer, shown in \cref{fig:hardware_and_mrvisu}B, supporting up to approximately \num{200e3} neurons and over \num{43e6} synapses.
Each neuron implements the \gls{adex} model~\cite{gerstner2009adex} with conductance-based synapses.
By linking multiple neuron circuits, a fan-in of up to \num{14e3} synapses can be realized per combined neuron.
Moreover, synapse circuits are plastic and provide pre-post-correlation observable data, supporting eg., spike-timing-dependent plasticity.
The neural network topology can be flexibly configured by a wafer-wide circuit switched network which transmits digital pulses in continuous time.

In a fully assembled \gls{bss1} system, as depicted in \cref{fig:hardware_and_mrvisu}A, a wafer connects to \num{48} \glspl{fpga} and various auxiliary \glspl{pcb}.
Connected to a local control cluster via a \SI{1}{GbE} link per \gls{fpga}, the system is accessible from conventional computing resources.
For a detailed description of the system, readers are referred to~\cite{schmidt2023commissioning}.

\section{Model Adaptation}

To illustrate the advantages of analog hardware, we investigate two biologically inspired network structures: the balanced random network described in~\cite{brunel_jcns2000} and the cortical microcircuit model detailed in~\cite{potjans2012cell}.

The structure of the balanced random network model is depicted in \cref{fig:hardware_and_mrvisu}D.
It comprises an inhibitory and an excitatory population of \gls{lif} neurons, which are randomly connected with a fixed connection probability of \num{0.1}.
Additionally, each neuron receives excitatory external input in the form of a Poisson-distributed stimulus with a rate $\nu_\mathrm{ext}$.
During the experiment, various configurations of this rate---defined relative to the cutoff frequency $\nu_\mathrm{thres}$, where neurons reach their threshold in the absence of internal stimulation---are examined, along with different inhibitory weights $w_\mathrm{inh}$ defined relative to the excitatory weights $w_\mathrm{exc}$.

In contrast, the cortical microcircuit model, illustrated in \cref{fig:hardware_and_mrvisu}E, includes four excitatory and four inhibitory populations of \gls{lif} neurons.
Their connectivity follows a predefined connectivity map, as detailed in~\cite{potjans2012cell}.
Each population receives a fixed number of excitatory external inputs, each modeled as a Poisson-distributed stimuli with a spike rate of \SI{8}{\hertz}.

Unlike digital simulators, analog hardware offers less flexibility in parameter and model configuration.
Consequently, before emulation on the \gls{bss1} system, the published models require adaptations to align with the hardware's specific constraints.

Experiments on \gls{bss1} are defined using PyNN~\cite{davison2009pynn}, a simulator-independent Python-based library for representing spiking neural network experiments, which is also supported by the NEST simulator~\cite{gewaltig2007nest}.
This compatibility facilitates the comparison of adapted models by allowing the adaptations to be incorporated directly into simulations.
Although these adapted models aim to replicate the first-order firing statistics of the original networks, achieving identical neuron behavior remains unfeasible.
Nonetheless, these models allow us to evaluate the advantages of analog neuromorphic hardware.

In this section, we present an overview of the necessary model modifications.
For a detailed discussion of these adaptations, please refer to~\cite{schmidt2024phd}.

Both models---the balanced random network and the cortical microcircuit---exceed the capacity of a single \gls{bss1} system, containing \num{12400} and \num{80000} neurons and \num{15625000} and \num{300000000} internal synaptic connections, respectively.
This limitation arises from the need to interconnect multiple neuron circuits to increase the fan-in, as well as the limited routing possibilities between neurons and synapses and the use of non-optimal mapping algorithms.
To make these models realizable on the hardware, we uniformly downscale both the neuron count and in-degree, maintaining the original connectivity probability.
To compensate for the reduced input resulting from this downscaling, the synaptic weights are linearly increased, following the approach outlined in~\cite{albada2015scaling}.

Even with this adjustment, some synaptic connections cannot be realized within the hardware constraints, leading to partial synapse loss distributed across all populations~\cite{petrovici2014characterization}.
This synapse loss is incorporated into the models, yielding scaled versions with \num{2083} neurons and \num{690157} internal synapses for the balanced random network, and \num{7712} neurons with \num{2373933} synapses for the cortical microcircuit.

Furthermore, as suggested in the original publication, the external Poisson input of the cortical microcircuit is replaced with an external current, implemented through an elevated leak potential.
In the balanced random network model, the external input is substituted by a pool of \num{2083} sources that generate Poisson-distributed spike trains, from which each internal neuron samples \num{200} connections.
This modification preserves the original stimulation strength, with a connectivity probability analogous to that of internal connections.

In addition, current-based synapses in the original models are converted to conductance-based synapses, with weights calculated to ensure that the amplitude of the post-synaptic potential, originating from the expected mean membrane potential, remains unchanged.
Finally, the short synaptic time constants are increased to align with hardware limitations, and normally distributed parameter variations are introduced across all neuron parameters estimated from calibration results of the hardware.

\section{Emulation on BrainScaleS-1}

By using models compatible with the \gls{bss1} system that retain the same connectivity complexity and firing statistics as their archetypes, we can demonstrate the unique features of the hardware through the emulation of these networks.
To this end, the models are mapped onto the hardware topology using the \gls{bss1} operating system as described in~\cite{mueller2022operating}.
This translation is required only once for each new network structure.
Subsequently, precalculated results can be loaded during hardware configuration.
An example of a routing result for the cortical microcircuit model is shown in~\cref{fig:hardware_and_mrvisu}C.

\begin{figure}
	\centering
	\includegraphics[width=\columnwidth]{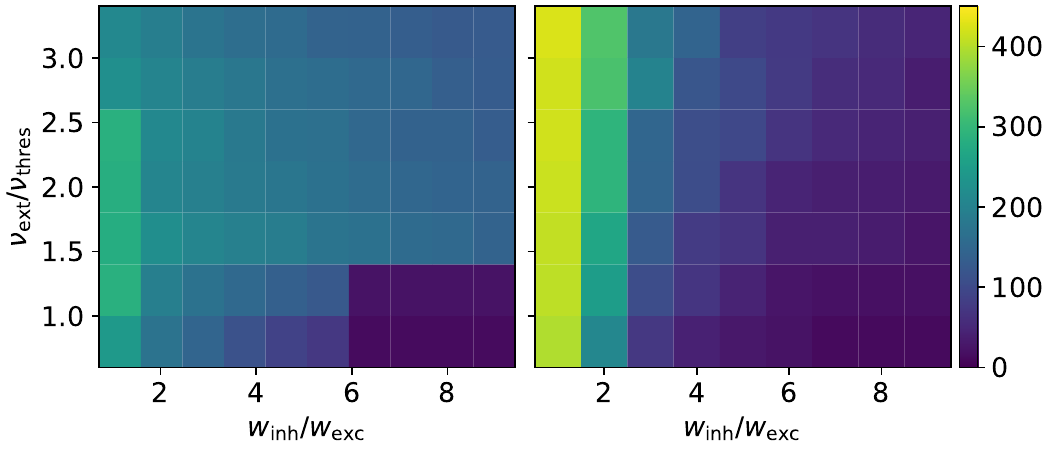}
	\caption{Mean firing rate of neurons in the balanced random network model, emulated on \gls{bss1} (left) and simulated in NEST (right).
		 Firing rates are reported across various relative inhibitory weights and external input spike rates.
		 For firing rates exceeding \SI{30}{\hertz}, saturation effects on the hardware introduce deviations in network behavior.}
	\label{fig:random_model_results}
\end{figure}

The firing behavior of neurons during the emulation of the balanced random network is presented in~\cref{fig:random_model_results}.
As in the original model, distinct firing regimes are observable.
However, additional hardware effects, such as saturating synaptic input circuits that are not accounted for in the adapted model, lead to deviations between simulation and emulation at biologically implausible mean firing rates above \SI{50}{\hertz}.
Despite these effects, the emulation successfully demonstrates the hardware's capability to internally route spikes at mean firing rates of up to \SI{250}{\hertz} within the biological domain.
Furthermore, by limiting spike readout to \num{30} neurons distributed across individual ASICs, we circumvent off-chip bandwidth limitations in high firing regimes with mean firing rates exceeding \SI{30}{\hertz}.
This approach ensures successful spike injection and readout, even with the large speed-up factor of \num{10000}.

Moreover, the emulation allows us to showcase the hardware's reconfigurability.
The initial loading of mapping results from disk and configuration of the hardware takes approximately one minute, but this step only needs to be performed once.
Subsequent adjustments, such as modifying external inputs, reprogramming digitally stored weights, and emulating the network for \SI{10}{\second} of biological time, can be completed in about \SI{10}{\second} of wall clock time.
Therefore, the hardware effectively achieves biological real-time performance even for relatively short emulation intervals, enabling efficient iterative executions.

\begin{figure}
	\centering
	\includegraphics[width=\columnwidth]{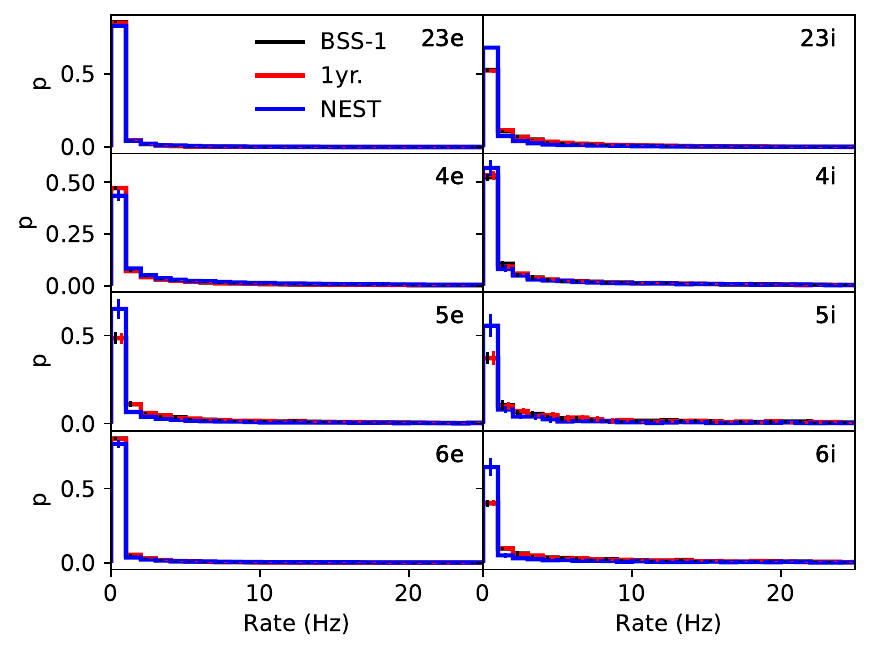}
	\caption{Firing rate distribution of neurons across the eight populations within the cortical microcircuit, emulated on \gls{bss1} and simulated in NEST.
	Results are extracted from a \SI{9}{\second} interval of biological time, starting \SI{1}{\second} after the experiment onset.
	Furthermore, the emulation is reevaluated after the network has evolved over a \num{1}-year period of biological time on the neuromorphic hardware.}
	\label{fig:microcircuit_results}
\end{figure}

Having demonstrated the system's usability, we can explore the advantages that become particularly evident in long-duration experiments, where the system's operational overhead becomes negligible.
In~\cref{fig:microcircuit_results}, the firing behavior of the cortical microcircuit is shown.
Although the model does not incorporate all hardware-specific features, the observed firing behavior remains comparable to the numerical simulation.
Notably, on the hardware, it is possible to reevaluate the network after \num{53} minutes, which corresponds to more than one year of biological network evolution---an achievement that is currently beyond the reach of conventional computing resources within a reasonable runtime.
While the network is expected and observed to remain stable, this example, combined with the capability for iterative weight updates and external stimulus injection, illustrates a clear advantage of the \gls{bss1} hardware.

\section{Discussion}

\begin{table}
\caption{Performance and energy comparison, based on the cortical microcircuit's network structure.
	For the BrainScaleS-1 system, the adapted model developed in this work is considered.\label{tab:performance_and_energy}}
\begin{tabular}{lS@{}S@{}}
\toprule
Simulator                                         & {Performance}          & {Energy} \\
                                                  & { (\num{e9} synaptic event/s) } & { (\si{\micro\joule}/synaptic event) } \\
\midrule
BrainScaleS-1                                     & 162                  & {\textless}0.012 \\
NeuroAIx-Framework$^{*}$ \cite{kauth2023neuroaix} & 19                   & 0.048 \\
CsNN$^{*}$ \cite{heittmann2022simulating}         & 3.8                  & 0.783 \\
NEST$^{*}$ \cite{kurth2022sub}                    & 1.8                  & 0.48 \\
SpiNNaker \cite{rhodes2020real}                   & 0.9                  & 0.6 \\
\bottomrule
\end{tabular}
\vspace{.25\baselineskip}
\\
\footnotesize{%
$^*$ Values are estimated from the reported speedup factor and the network behavior of the full-scale model with external Poisson inputs.
}\\
\end{table}

Having emulated two networks with biologically inspired connectivity, we discuss the performance of the hardware.
Especially the cortical microcircuit, with its complex recurrent connectivity represents a common benchmark in computational neuroscience~\cite{ostrau2022}.
Based on its network structure, \cref{tab:performance_and_energy} provides a comparative analysis of performance and energy for various simulators and hardware backends.
The energy estimation of the \gls{bss1} system is calculated assuming the worst-case power consumption of \SI{2}{\kilo\watt} for the entire system.
However, we assume the actual power consumption to be considerably lower.

Although our investigated model differs in size, no substantial efficiency gains are anticipated across networks of varying sizes with similar structures.
For smaller networks, simulation speed is limited by the minimal number of dependent numerical instructions and their execution latency, whereas for larger networks it is limited by communicating spike signals among inter-connected neurons~\cite{kauth2023neuroaix}.
Accordingly, the presented results on \gls{bss1} demonstrate, to the best of our knowledge, the fastest operation to date in terms of the number of synaptic events per second in a network exhibiting the complexity of the cortical microcircuit.
It should be noted that the observed performance is a consequence of the large speedup factor of \num{10000} that is inherent to the emulation, which remains independent of the implemented network size.
The operational overhead that is introduced by the configuration of the system and the transmission of recorded observables (spikes) is not incorporated into the presented results.
Consequently, the potential for exploiting the full speedup of the system is most evident for long or repetitive emulations.
Nevertheless, in combination with the comparably low energy consumption during emulation, we illustrated the advantages of the physical modeling approach of the wafer-scale substrate.

The models evaluated in this work do not include any form of plasticity; they are therefore relatively inexpensive to compute numerically.
In applications where, e.g., synaptic plasticity is used, it is expected that the advantage of physical model emulation over numerical simulation will be significantly greater.

Furthermore, we want to emphasize that the \gls{bss1} system is based on \SI{180}{\nano\meter} CMOS technology (introduced in 1999).
In comparison, recent architectures, such as those described in~\cite{kauth2023neuroaix} and~\cite{heittmann2022simulating}, utilize smaller node sizes of \SI{28}{\nano\meter} (introduced in 2010).
Therefore, future neuromorphic systems leveraging these smaller nodes could incorporate more versatile circuits, enhancing experimental flexibility.
To a certain extent, the current \gls{bss2} architecture based on \SI{65}{\nano\meter} (introduced in 2005) already offers significant improvements~\cite{pehle2022brainscales2_nopreprint_nourl}, although the upscaling of the platform is still in progress.

Neuromorphic hardware will always come at a cost:
New network architectures must be specifically mapped to the hardware, and each experiment requires initial system configuration.
Thus, we expect the optimal use of these systems to be in conjunction with conventional simulators.
In this co-execution approach, diverse network topologies can be explored in simulation, while the neuromorphic hardware handles continuous time emulation, extended-duration experiments, and iterative parameter sweeps, benefiting from its fast dynamical evolution that is independent of network size and model complexity.

The open research infrastructure EBRAINS\footnote{\url{https://www.ebrains.eu/}} further streamlines this integration.
There, networks can be designed using the PyNN description language and then directed either to HPC clusters for simulation or to the \gls{bss1} system for hardware-based emulation, allowing researchers worldwide to run experiments remotely.
Both experiments presented in this paper are available and executable through EBRAINS\footnote{Available at \url{https://github.com/electronicvisions/brainscales1-demos}}, promoting accessibility and reproducibility.

With these advancements, neuromorphic hardware provides a powerful complement to traditional simulation methods, offering new avenues for computational neuroscience.

\section*{Acknowledgments}
\label{sec:acknowledgements}
The authors wish to thank all present and former members of the Electronic Vision(s) research group contributing to the \acrlong{bss1} neuromorphic platform.

\noindent%
\subsection*{Funding}
This work has received funding from
the EC Horizon 2020 Framework Programme
under grant agreements
785907 (HBP SGA2) 
and
945539 (HBP SGA3), 
the EC Horizon Europe Framework Programme
under grant agreement
101147319 (EBRAINS 2.0),
the \foreignlanguage{ngerman}{Deutsche Forschungsgemeinschaft} (DFG, German Research Foundation) under Germany’s Excellence Strategy EXC 2181/1-390900948 (the Heidelberg STRUCTURES Excellence Cluster),
and the Helmholtz Association Initiative and Networking Fund [Advanced Computing Architectures (ACA)] under Project SO-092.

\printbibliography

\end{document}